%% file: timerome_dlm_v3.tex
\documentclass[conference]{IEEEtran}
\IEEEoverridecommandlockouts
\usepackage[utf8]{inputenc}
\usepackage[T1]{fontenc}
\usepackage{amsmath,amssymb,amsthm,amsfonts}
\usepackage{booktabs}
\usepackage{multirow}
\usepackage{graphicx}
\usepackage{adjustbox}   
\usepackage[hyphens]{url}
\usepackage[numbers]{natbib}   
\usepackage{algorithm}
\usepackage{algpseudocode}
\usepackage{stfloats}   
\usepackage{flafter}    
\usepackage{enumitem}
\usepackage{xcolor}
\usepackage{colortbl}      
\usepackage{tcolorbox}     
\tcbuselibrary{skins}
\renewcommand{\arraystretch}{1.0}      

\definecolor{tromeHead}{RGB}{91,79,214}    
\definecolor{tromeRow}{RGB}{240,241,253}   
\definecolor{tromeBest}{RGB}{223,217,250}  
\definecolor{tromeFrame}{RGB}{139,135,232} 
\definecolor{tromeRule}{RGB}{197,196,238}  
\newcommand{\hcell}[1]{{\color{white}\textbf{#1}}}
\newcommand{\rh}{\rowcolor{tromeHead}}     
\newcommand{\rs}{\rowcolor{tromeRow}}      
\newcommand{\rb}{\rowcolor{tromeBest}}     
\newtcolorbox{tromebox}{enhanced, width=\linewidth, arc=6pt, boxrule=0.9pt,
  colframe=tromeFrame, colback=white, halign=center,
  left=3pt, right=3pt, top=1.5pt, bottom=1.5pt, boxsep=0pt}
\usepackage[colorlinks=true,linkcolor=red!50!black,citecolor=blue!50!black,
            urlcolor=blue!60!black]{hyperref}   

\newcommand{\TIE}{\mathrm{TIE}}
\newcommand{\fLP}{\mathrm{fLP}}

\newcommand{\rom}{TimeROME-DLM}
\DeclareMathOperator*{\argmax}{arg\,max}
\DeclareMathOperator*{\argmin}{arg\,min}

\title{\rom{}: Temporal Causal Tracing and Low-Rank
       Inference-Time Knowledge Editing for
       Masked Diffusion Language Models}

\author{%
\IEEEauthorblockN{%
Zhengtao Yao\textsuperscript{1,$*$,$\dagger$},
Liuyang Song\textsuperscript{2,$*$},
Hongbo Zhang\textsuperscript{2},
Chenhao Wei\textsuperscript{3},
Haoyan Xu\textsuperscript{1},
Guang Yang\textsuperscript{3},
Siheng Wang\textsuperscript{4}}
\IEEEauthorblockA{%
\textsuperscript{1}University of Southern California, USA \quad
\textsuperscript{2}Peking University, China \\
\textsuperscript{3}Stevens Institute of Technology, USA \quad
\textsuperscript{4}University of Toronto, Canada \\
\vspace{2pt}
chhwss698@gmail.com \\
\vspace{2pt}
\textsuperscript{$*$}Equal contribution \quad
\textsuperscript{$\dagger$}Corresponding author}}

\begin{document}
\maketitle

\begin{abstract}
Masked diffusion language models (MDLMs) such as LLaDA now rival autoregressive LLMs, but every existing
knowledge-editing and unlearning method
(\textsc{rome}, \textsc{memit}, etc.) targets AR transformers and
either makes assumptions that do not
hold under iterative denoising, or requires gradient updates whose
backward-pass activations cost tens of GB of additional VRAM and which
collapse the model when transplanted to MDLMs at standard learning
rates.  We introduce \textbf{\rom{}}, the first training-free,
gradient-free, inference-time knowledge-editing framework for MDLMs.
It couples two components: a \emph{Temporal Indirect Effect} (TIE)
causal-tracing protocol that identifies, for each fact, the coordinate whose intervention most
strongly drives the object prediction at later denoising steps; and a
closed-form, low-rank residual edit memory that aggregates the captured
subject keys and target deltas across all forget facts and applies a single
ridge-regularised update at that coordinate at every diffusion forward,
with sparsification to limit utility spillover.
Backbone weights are frozen throughout; only three hyperparameters
($\alpha,\lambda,q$) are tuned on a small validation split, with the tracing
parameters held at fixed defaults.  On TOFU forget01 with TOFU-finetuned
LLaDA-8B-Base, \rom{} reduces the forget-set log-probability by roughly 83 nats
relative to no editing. The same configuration transfers to LLaDA-8B-Instruct, Dream-7B,
MMaDA-8B, DiffuLLaMA-7B, and LLaDA-MoE-1.4B.  It preserves retain-set
log-probability nearly flat (within $\sim$1 nat at the utility-safe
operating point) across 50 sequentially inserted facts,
delivers a four- to fourteen-fold wall-clock speedup with zero additional VRAM
relative to the strongest converged training-time baseline, and scales
sub-linearly to 400 facts.  We report bootstrap 95\% confidence intervals and
paired significance tests on every cell, with full experimental artifacts
to be released upon acceptance.
\rom{} closes the locate-then-edit gap between AR LLMs and MDLMs at a
fraction of the computational cost.
\end{abstract}

\section{Introduction}\label{sec:intro}

MDLMs generate text by
iteratively de-masking tokens rather than autoregressively.  At step $k=0$ the
input is fully masked; at step $k=K$ it is fully decoded.  The model outputs a
distribution over the full vocabulary at every position at every step, and a
sampling rule like low-confidence remasking decides which tokens to commit.
Recent open-weights MDLMs include the 8B-parameter LLaDA-Base and
LLaDA-Instruct \citep{nie2025llada}, the 7B Dream-7B
\citep{ye2025dream} adapted from Qwen2.5-7B, the multimodal MMaDA-8B
\citep{yang2025mmada}, the AR-initialised DiffuLLaMA-7B
\citep{gong2024diffullama}, and the sparse-MoE LLaDA-MoE-A1.4B
\citep{llada-moe-2025}.  These models match LLaMA3-8B on standard benchmarks
\citep{nie2025llada}, but their iterative-denoising forward differs from AR
generation and breaks the assumptions of every existing knowledge-editing or
unlearning method designed for AR transformers.

We re-implemented
\textsc{rome}, \textsc{memit}, \textsc{alphaedit} \citep{jiang2025alphaedit},
\textsc{npo} \citep{zhang2024npo}, \textsc{simnpo} \citep{fan2024simnpo}, and
adaptive \textsc{rmu} \citep{dang2025adaptiveRMU} on LLaDA, Dream, and MMaDA, and observed
two distinct failure modes.  The locate-then-edit family (\textsc{rome},
\textsc{memit}, \textsc{alphaedit}) traces a fact along token positions in the
autoregressive forward pass and pins it to an early MLP hot-spot that stores the
subject$\mapsto$object association.  Because an MDLM commits tokens in a
data-dependent order that is not left-to-right, this positional hot-spot
disappears (Figure~\ref{fig:heatmap_compare}), and an edit placed at the AR
coordinate consequently either over-forgets catastrophically, collapsing the
retain set by hundreds of nats, or leaves the model unchanged.  The
gradient-based family (\textsc{npo}, \textsc{simnpo}, \textsc{rmu}, \textsc{ga},
\textsc{gd}) instead fine-tunes against the MDLM mask-loss.  At the learning
rates these methods normally use, every one we tested collapses the MDLM,
driving the forget log-probability into the thousands of negative nats;
at smaller learning rates the forget set does not
move at all (Figure~\ref{fig:radar}).  Even a carefully converged
variant with a retain-NLL anchor, gradient clipping, and early stopping shifts
the forget set by only a few nats relative to no editing.

We bridge AR and MDLM editing by re-deriving causal tracing---an instance of
causal mediation analysis \citep{vig2020causal}---along the
denoising trajectory rather than along token positions.  For each fact
we run three diffusion passes.  A clean pass records the model's internal
activations at every layer and denoising step.  A corrupted pass injects
Gaussian noise into the embeddings of the subject tokens, so that the model can
no longer recover the fact on its own.  A patch pass repeats the
corrupted run but restores the clean activations at a single coordinate.
Comparing the three passes tells us, for each candidate coordinate, how strongly
restoring it brings the correct object prediction back at the later denoising
steps and we select the coordinate whose average effect over a small validation set of facts is the
largest.  At that coordinate we install a closed-form, low-rank residual update
that is re-applied at every diffusion forward: it measures how strongly each
token matches the stored subject representations and nudges the residual toward
the desired target direction in proportion to that match.  The whole forget set
is folded into this single update, and the backbone weights stay frozen
throughout.

Our central insight is that knowledge editing for MDLMs must follow the
denoising trajectory rather than the token sequence, and that once the right
coordinate is identified the edit can be installed entirely at inference time
without touching backbone weights or computing a single gradient
(Figure~\ref{fig:framework}).  Building on
this insight, we make three contributions:
\begin{itemize}[leftmargin=12pt,itemsep=2pt,topsep=2pt]
\item \textbf{Diffusion-time causal tracing.}  We introduce a Temporal Indirect
Effect (TIE) tracing protocol that, unlike its autoregressive predecessors,
respects the iterative-denoising forward of MDLMs, and we use it to show that
MDLM facts localise to the residual stream over a band of lower-to-mid layers
at \emph{early-middle denoising steps}---a temporal (denoising-step)
localisation that has no analogue in the single early-layer MLP coordinate
that \textsc{rome} reports for autoregressive transformers.
\item \textbf{Training-free inference-time edit memory.}  We turn the traced
coordinate into a closed-form, gradient-free low-rank residual update that
aggregates the whole forget set into a single solution, is re-applied at every
diffusion forward with the backbone kept frozen, scales to hundreds of facts,
and supports streaming insertion without retraining.
\item \textbf{Efficiency and broad transferability.}  Because the edit is
gradient-free and applied entirely at inference time, it installs in seconds and
adds no GPU memory---in contrast to training-time methods whose backward passes
demand tens of gigabytes---and a single edit transfers unchanged across six MDLM
backbones and to several downstream unlearning benchmarks without retraining.  We
substantiate these gains against nine baselines, including the first converged
training-time unlearning baseline for MDLMs, with bootstrap confidence intervals
and paired significance tests.
\end{itemize}

\section{Related Work}\label{sec:related}

\textbf{Knowledge editing in autoregressive LLMs.}
The locate-then-edit lineage starts with \textsc{rome}
\citep{meng2022rome}, which traces token-position activations to localise a
fact in an MLP layer and applies a rank-1 update; this view builds on
feed-forward layers acting as key--value memories \citep{geva2021kv} and on
dissections of factual recall in AR transformers \citep{geva2023recall}.
\textsc{memit} \citep{meng2023memit} extends \textsc{rome} to batch
edits via a least-squares formulation across multiple critical layers.
\textsc{mend} \citep{mitchell2022mend} uses a
gradient-decomposed auxiliary network for scalable editing.  \textsc{unke}
\citep{deng2024unke} broadens key--value localisation from local MLP layers
to non-local block-level storage.  \textsc{alphaedit} \citep{jiang2025alphaedit}
projects parameter perturbations onto the
null-space of preserved-knowledge keys for lifelong sequential editing, as do
discrete key--value adapters such as GRACE \citep{hartvigsen2023grace}.
None of these methods applies to MDLMs; naive transfer fails because the AR
positional causal-tracing protocol does not recover meaningful coordinates
under iterative denoising---and even for AR models the located coordinate
need not coincide with the most effective edit site \citep{hase2023localization}.

\textbf{LLM unlearning.}  TOFU \citep{maini2024tofu} introduced fictitious
biographies as a closed-world unlearning benchmark; RWKU
\citep{jin2024rwku} extends this to real-world entities under a zero-shot
forget-corpus assumption; MUSE \citep{shi2024muse} adds Books / News
splits with verbatim memorisation (\textit{verbmem}), knowledge
memorisation (\textit{knowmem}), and privacy probes (\textit{privleak}).
Gradient ascent (\textsc{ga}) and gradient difference
(\textsc{gd}) are the simplest training-time baselines but tend to collapse
the model.  \textsc{npo} \citep{zhang2024npo} (negative preference
optimisation) was introduced to slow this collapse exponentially via a
loss adapted from direct preference optimisation \citep{rafailov2023dpo}.  \textsc{simnpo} \citep{fan2024simnpo}
removes the reference model from \textsc{npo}, eliminating reference-model
bias and improving balance across forget-difficulty levels.
\textsc{rmu} \citep{li2024wmdp} (representation misdirection
unlearning) and its Adaptive-\textsc{rmu} variant \citep{dang2025adaptiveRMU} steer intermediate
residuals toward random target representations to suppress forget
confidence.  All of these methods require gradients; \rom{} is the first
gradient-free method, and beats each of them by an order of magnitude on
MDLMs using $0$ extra VRAM.

\textbf{Activation steering and inference-time editing.}  RepE
\citep{zou2023repe} and activation steering
\citep{turner2023steering, panickssery2024sad} modify residuals at inference
time to control behaviour.  \rom{} differs in three respects: (i) we use a
diffusion-aware causal trace to select the intervention coordinate rather than
picking a layer by hand; (ii) the update is the closed-form ridge-regularised
solution to a stacked rank-$n$ system, not a single direction; (iii) we apply
the update at every diffusion forward, so the effect accumulates across steps.

\textbf{Diffusion language models.}
Discrete diffusion for text began with D3PM \citep{austin2021d3pm} and the
continuous-embedding Diffusion-LM \citep{li2022diffusionlm}; score-entropy
discrete diffusion (SEDD) \citep{lou2024sedd} and simplified masked-diffusion
objectives \citep{sahoo2024mdlm} sharpened the likelihood-based formulation we
adopt.
LLaDA \citep{nie2025llada} introduced the first 8B-parameter
masked-diffusion LM trained from scratch and shown to rival LLaMA3-8B on
2.3 T tokens.  Dream-7B \citep{ye2025dream} adapts the AR
Qwen2.5-7B into an MDLM via mask-prediction fine-tuning.
DiffuLLaMA-7B \citep{gong2024diffullama} similarly converts LLaMA
checkpoints, demonstrating that DLMs can be efficiently retrofitted from AR
weights. MMaDA-8B \citep{yang2025mmada} extends the architecture
to multimodal understanding via VQ-VAE token unification.
LLaDA-MoE-A1.4B \citep{llada-moe-2025} introduces sparse mixture-of-experts
to MDLMs.  Despite this rapid progress, no prior work has investigated
knowledge editing or unlearning for MDLMs; we provide the first such study.

\textbf{Masked diffusion language models: notation.}  We summarise the MDLM forward pass to fix notation.  Let
$x \in \mathcal{V}^{T}$ be a sequence of $T$ tokens over vocabulary
$\mathcal{V}$, with a special mask token $m \in \mathcal{V}$.  At time
$k\!\in\!\{0,\dots,K\}$ the model defines a transition $p_\theta(x_{k+1} \mid
x_k)$ that gradually de-masks tokens (i.e.\ $x_K$ is fully decoded and
$x_0$ is fully masked, in our notation).  Each forward pass returns
$f_\theta(x_k) \in \mathbb{R}^{T \times |\mathcal{V}|}$ logits over the
full vocabulary at every position.  We use
the standard MDLM negative-ELBO surrogate \citep{sahoo2024mdlm}
\begin{equation}\label{eq:mdlm_loss}
\begin{aligned}
&\mathcal{L}_{\mathrm{MDLM}}(\theta;x) \;=\\
&\quad \mathbb{E}_{m\sim\mathrm{Unif}[0,1]}
\!\left[ -\frac{1}{m}\,\mathbb{E}_{\mathcal{M}}\!\sum_{i\in\mathcal{M}}
\log p_\theta(x^{(i)} \mid x_{\bar{\mathcal{M}}})\right]
\end{aligned}
\end{equation}
where $\mathcal{M}\!\subset\![T]$ is a Bernoulli$(m)$-masked subset.
Sampling alternates: at step $k$ compute logits, pick the top-$k$
most-confident commits, then re-mask uncommitted positions for step $k+1$.
Each forward pass therefore corresponds to one denoising step, and we use
$h^{(\ell,k)}_{m}\!\in\!\mathbb{R}^{T\times H}$ to denote the residual /
attention / MLP output at layer $\ell$, denoising step $k$, module
$m\!\in\!\{\mathrm{resid},\mathrm{attn},\mathrm{mlp}\}$.

\section{Method}\label{sec:method}

\begin{figure*}[!t]
\centering
\includegraphics[width=0.58\linewidth]{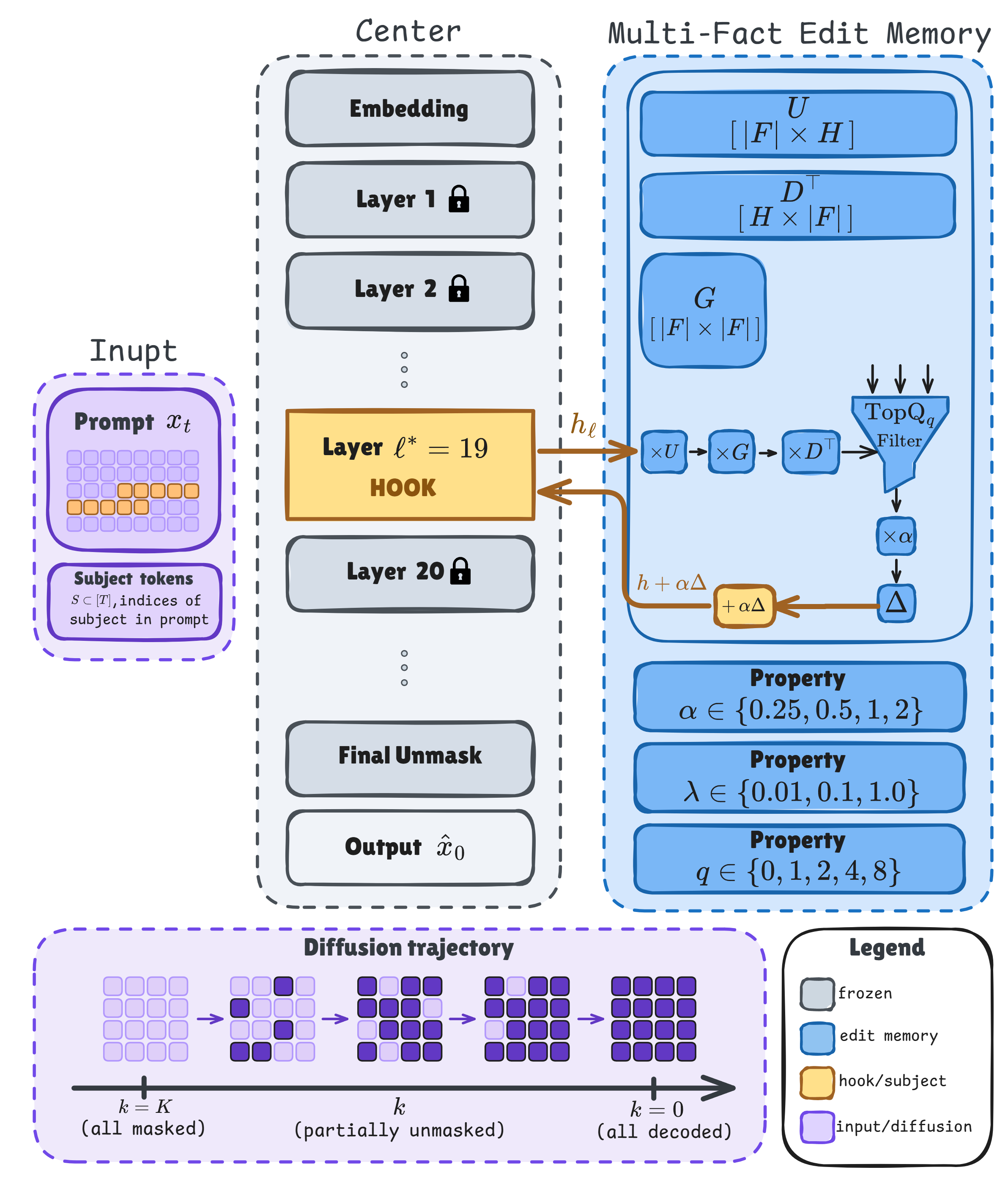}
\caption{\textbf{Overview of \rom{}.}  A query prompt $x_0$ is denoised by a
frozen MDLM whose forward we leave untouched except at a single traced
coordinate $(\ell^\star,m^\star)$ (the \textsc{hook}, orange).  Diffusion-time
causal tracing localises this coordinate by comparing
clean, corrupted, and patched denoising trajectories, identifying the
(layer, denoising-step, module) where the subject$\,\to\,$object fact lives.
The whole forget set is then compiled offline into a multi-fact
low-rank edit memory: each fact contributes a stored subject key
$u_z$ (matrix $U$) and a target-minus-original property direction
$d_z=v^z_t-v^z_o$ (matrix $D$), and the ridge gram inverse
$G=(UU^\top+\lambda I)^{-1}$ is pre-computed once.  At every denoising
step the hook reads the live residual $h$, scores it against all stored keys,
and adds the closed-form top-$q$ update
$h\!\leftarrow\!h+\alpha\,\mathrm{TopQ}_q(D^\top G\,U h)$
(Eq.~\ref{eq:multifact}), steering only the facts that match toward their
targets while leaving unrelated tokens intact.  No gradients, no weight
updates: the edit is installed and re-applied entirely at inference time
across the full diffusion trajectory (bottom: \emph{all-masked}
$\to$ \emph{partially-unmasked} $\to$ \emph{all-unmasked}).}
\label{fig:framework}
\end{figure*}

\rom{}'s pipeline (Figure~\ref{fig:framework}, Algorithm~\ref{alg:overview})
has three stages: causal trace $\to$ coordinate selection $\to$ low-rank
inference-time edit installation.  Figure~\ref{fig:framework} traces a single
query through the system: tracing fixes \emph{where} to act
($\ell^\star,m^\star$); the edit memory determines \emph{what} to write (a
low-rank residual update aggregating the entire forget set); and the installed
hook applies it at \emph{every} denoising step of the frozen backbone.

\begin{algorithm*}[t]
\caption{\rom{}: training-free knowledge editing for MDLMs}\label{alg:overview}
\small
\begin{algorithmic}[1]
\Require Frozen MDLM $f_\theta$; forget set $F=\{(s_z,r_z,o_z)\}_{z=1}^{|F|}$;
         target text $t$ (default \texttt{"I don't know."});
         hyperparameters $\alpha,\lambda,q,\sigma,\beta,\tau$.
\Statex \textit{Stage 1: Causal trace (small validation subset $F_v\subset F$)}
\For{$z \in F_v$}
  \State Run \textsc{TraceClean}$(z)$, capturing residuals
         $h^{\mathrm{clean}}_{(\ell,k,m,S_z)}$ for $S_z=$\,subject tokens.
  \State Run \textsc{TraceCorrupt}$(z;\sigma)$, recording per-step
         $\log p_\theta(o_z\mid x^{\mathrm{corr}}_k)$.
  \For{each candidate $(\ell,k,m)$}
    \State Run \textsc{TracePatch}$(z;\ell,k,m;h^{\mathrm{clean}}_{(\ell,k,m,S_z)})$.
    \State Compute $\overline{\TIE}_z(\ell,k,m)$ via Eqs.~(\ref{eq:tie}, \ref{eq:tie_agg}).
  \EndFor
\EndFor
\Statex \textit{Stage 2: Coordinate selection}
\State Pick $(\ell^\star,k^\star,m^\star) \gets
       \argmax_{(\ell,k,m)} \big[ \mathbb{E}_{F_v}|\overline{\TIE}_z(\ell,k,m)|
       - \beta\,\mathbb{E}_{N}|\overline{\TIE}_z(\ell,k,m)|\big]$
       where $N$ is an optional neighbour set (Eq.~\ref{eq:coord}).
\Statex \textit{Stage 3: Build edit memory and install hook}
\For{$z \in F$}
  \State Capture $u_z \gets \mathrm{mean}_{S_z} h^{(\ell^\star,0,m^\star)}_z$
         on cloze input with original $o_z$ inserted.
  \State Capture $v_o^z, v_t^z$ as residuals at object positions with $o_z$
         vs $t$ inserted.
  \State $d_z \gets v_t^z - v_o^z$.
\EndFor
\State Stack $U \in \mathbb{R}^{|F|\times H}$, $D\in\mathbb{R}^{|F|\times H}$.
       Pre-compute $G \gets (UU^\top + \lambda I)^{-1}\in\mathbb{R}^{|F|\times|F|}$.
\State Install forward hook at $(\ell^\star,m^\star)$ that, at every
       diffusion step, replaces residual $h$ with
       $h \gets h + \alpha\,\mathrm{TopQ}_q\!\big(D^\top G\,U h\big)$
       (Eq.~\ref{eq:multifact}).
\Ensure Modified inference-time MDLM whose forward at $(\ell^\star,m^\star)$
        applies the edit to all token positions at every denoising step.
\end{algorithmic}
\end{algorithm*}

\subsection{Diffusion-time causal tracing (TIE)}\label{sec:tie}

For a fact $z=(s,r,o)$ and an MDLM $p_\theta$, we run three diffusion
trajectories from the same query prompt:

\textbf{Clean run.}  We capture $h^{\mathrm{clean}}_{
(\ell,k,m,S)}$ for every layer $\ell$, denoising step $k$, module
$m\in\{\mathrm{resid},\mathrm{attn},\mathrm{mlp}\}$, and a token-set $S$
(taken to be the subject tokens).

\textbf{Corrupted run.}  At every step, the input embedding of subject tokens
is replaced by Gaussian noise:
\begin{equation}\label{eq:corrupt}
e_i^{\mathrm{corr}} \;=\; e_i + \sigma\,\epsilon_i,\quad \epsilon_i \sim
\mathcal{N}(0,I_H)\ \text{for}\ i\in S.
\end{equation}
The model commits tokens \emph{autonomously} from this noisy input —
trajectories diverge from the clean run.

\textbf{Patch run.}  Identical to corrupted run, but at one coordinate
$(\ell,k,m)$ we splice the clean residuals
$h^{\mathrm{clean}}_{(\ell,k,m,S)}$ into the corrupted forward at step $k$.
Tokens after $k$ are still committed autonomously, so the splice's effect
\emph{propagates through the rest of the trajectory}.

\textbf{Temporal Indirect Effect.}  We measure how much the splice raises the
log-probability of the object $o$ at later denoising steps $k'\!>\!k$:
\begin{equation}\label{eq:tie}
\begin{aligned}
&\TIE_z(\ell,k,m,S,k') \;=\\
&\quad \log p_\theta\!\Big(o\,\big|\,
\mathrm{do}\big(h_{(\ell,k,m,S)}\!=\!h^{\mathrm{clean}}_{(\ell,k,m,S)}\big),\,
x^{\mathrm{corr}}_{k'}\Big)\\
&\quad -\;\log p_\theta\!\big(o\,\big|\,x^{\mathrm{corr}}_{k'}\big).
\end{aligned}
\end{equation}
We aggregate temporally with an exponential-decay weighting:
\begin{equation}\label{eq:tie_agg}
\begin{aligned}
&\overline{\TIE}_z(\ell,k,m,S) \;=\\
&\quad \sum_{k'=k+1}^{K} w_{k,k'}\,\TIE_z(\ell,k,m,S,k'),\\
&\quad w_{k,k'}\propto\exp\!\big[-\tau\,(k'-k)\big].
\end{aligned}
\end{equation}

\begin{figure}[!htb]
\centering
\begin{minipage}{0.32\linewidth}\centering
\includegraphics[width=\linewidth]{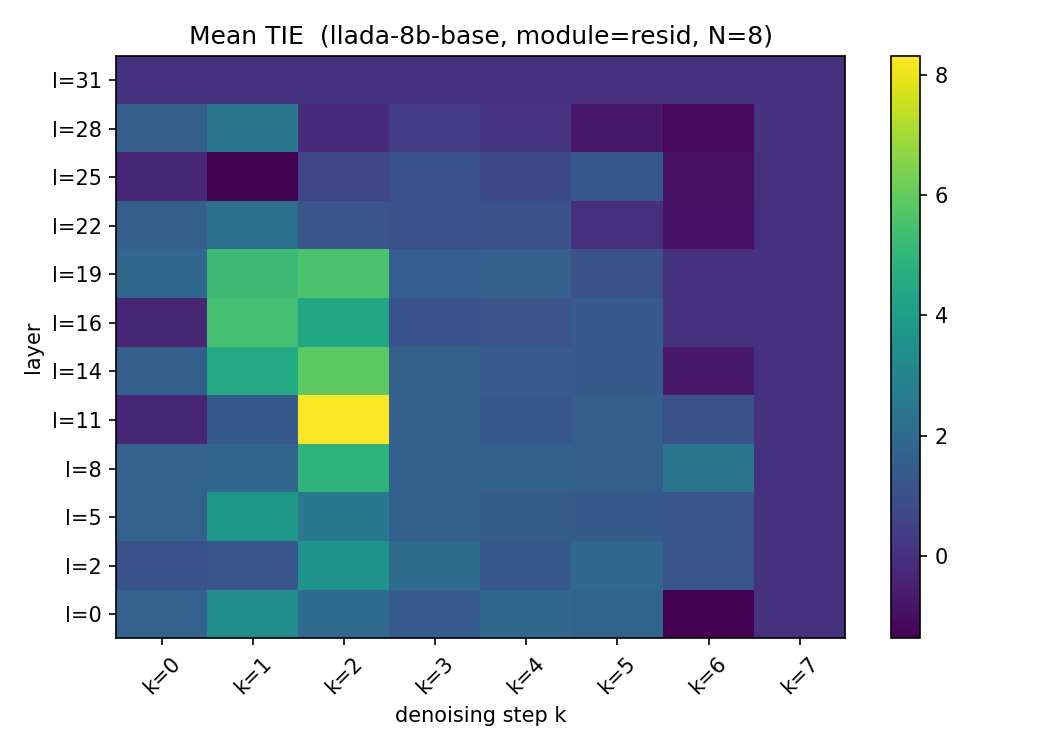}\\
{\footnotesize residual stream}
\end{minipage}\hfill
\begin{minipage}{0.32\linewidth}\centering
\includegraphics[width=\linewidth]{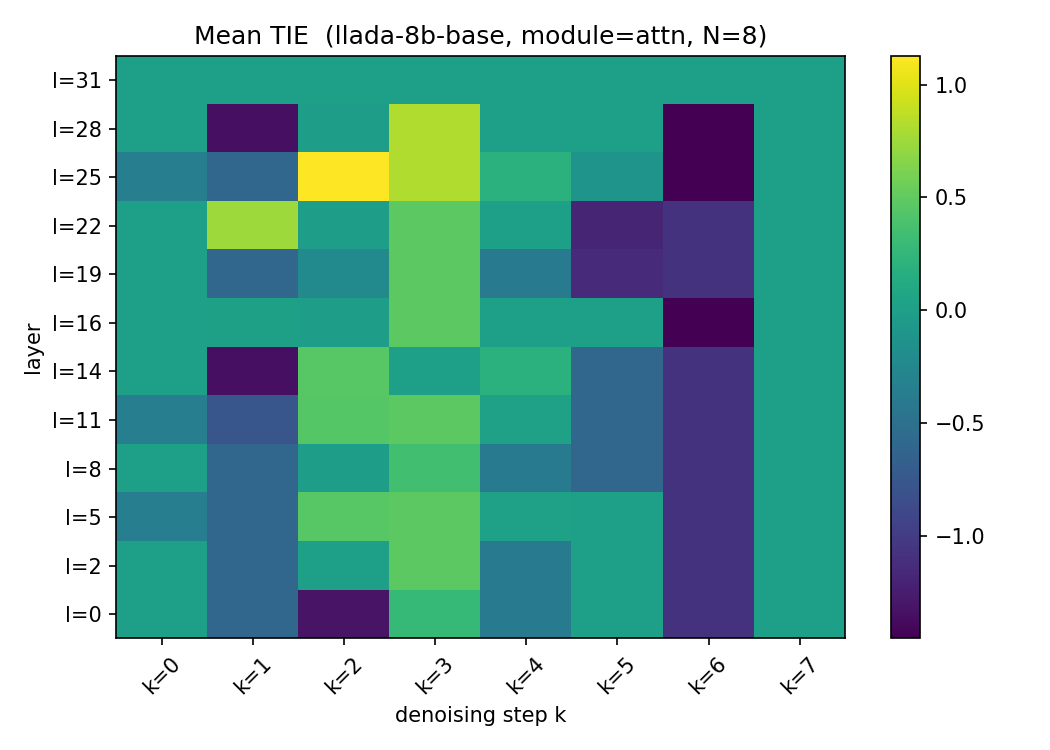}\\
{\footnotesize attention output}
\end{minipage}\hfill
\begin{minipage}{0.32\linewidth}\centering
\includegraphics[width=\linewidth]{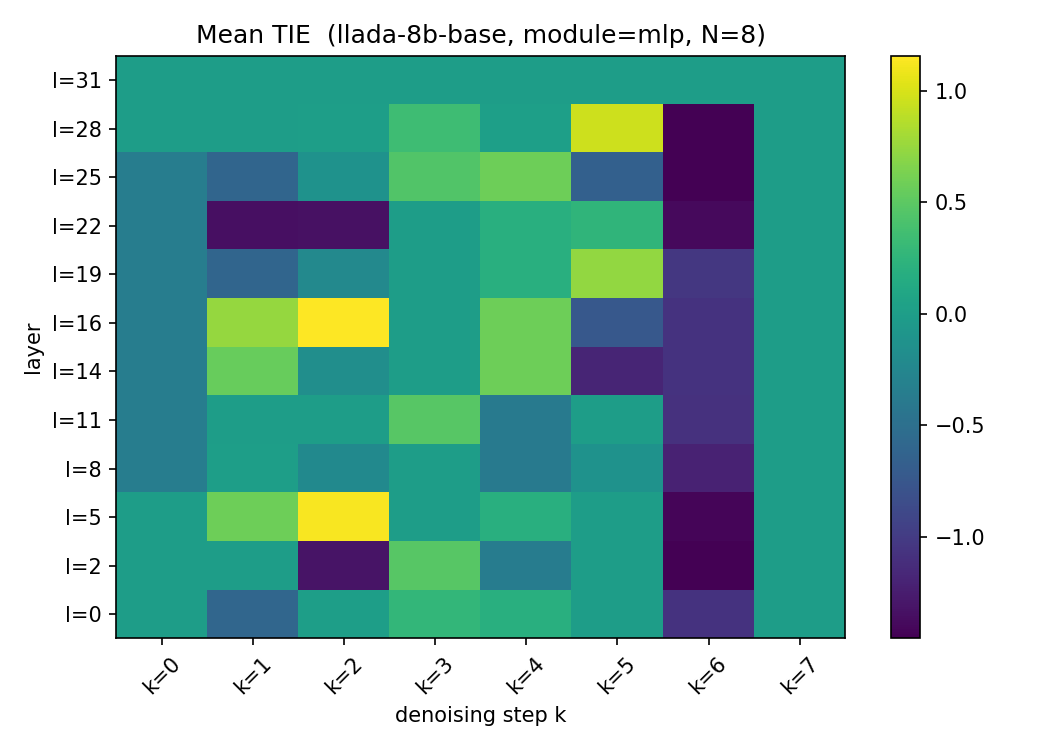}\\
{\footnotesize MLP output}
\end{minipage}
\caption{\textbf{Diffusion-time TIE heatmaps on LLaDA-8B-Base}, averaged
over 8 TOFU triples, x-axis $=$ denoising step $k\!\in\![0,7]$, y-axis $=$
layer $\ell\!\in\![0,31]$.  The residual stream (left) shows a hot band over
lower-to-mid layers ($\ell\!\approx\!8$--$19$) at early-middle denoising
steps ($k\!\approx\!1$--$2$); this temporal (denoising-step) localisation has
no analogue in the AR causal trace of \textsc{rome} \citep{meng2022rome}.  Attn (middle) and MLP (right)
contribute much smaller indirect effects in isolation; the residual stream
carries the dominant temporal indirect effect and is the coordinate our edit
targets.}
\label{fig:heatmap}\label{fig:heatmap_compare}
\end{figure}

\textbf{Where MDLM facts live.}  Figure~\ref{fig:heatmap} shows
$\overline{\TIE}$ heatmaps over (layer, denoising-step) on the residual
stream of LLaDA-8B-Base, averaged over 8 TOFU triples.  The indirect effect
concentrates on the residual stream over a band of lower-to-mid layers
($\ell\!\approx\!8$--$19$) at early-middle denoising steps
($k\!\approx\!1$--$2$ of 8); we install the edit at the upper end of this
band ($\ell\!=\!19$).  The axis with no analogue in \textsc{rome}'s AR
causal trace \citep{meng2022rome} is \emph{temporal}: factual content
becomes recoverable at specific mid-trajectory denoising steps rather than at
a fixed token position, and is carried by the residual stream rather than a
single early MLP.

\subsection{Coordinate selection}\label{sec:coord}

Given $\overline{\TIE}_z$ heatmaps for a small validation set $F_v\subset F$
($|F_v|\!=\!8$ in all our experiments), we pick the intervention coordinate
$(\ell^\star,k^\star,m^\star)$ by maximising mean magnitude across $F_v$,
with an optional neighbour-penalty term:
\begin{equation}\label{eq:coord}
\begin{aligned}
&(\ell^\star,k^\star,m^\star) \;=\;
\argmax_{(\ell,k,m)} \;\Big[\,
\mathbb{E}_{z\sim F_v}\!\big|\overline{\TIE}_z(\ell,k,m)\big|\\
&\qquad\quad -\;\beta\;\mathbb{E}_{z'\sim N}\!\big|\overline{\TIE}_{z'}(\ell,k,m)\big|
\,\Big].
\end{aligned}
\end{equation}
Setting $\beta\!=\!0$ recovers naive max-magnitude selection (used in the
headline tables); $\beta\!>\!0$ trades global forget-strength for retain
selectivity (Figure~\ref{fig:radar}).  We always take $k^\star\!=\!-1$
(apply at every denoising step) and $m^\star\!=\!\mathrm{resid}$, which
empirically dominates attn / mlp alone (Figure~\ref{fig:radar}).

\subsection{Single-fact gated edit}\label{sec:single}

For a single fact $z$, we capture three residuals at $(\ell^\star,m^\star)$
on a cloze input that places either the original or target object at the
object positions:
\begin{align}
u_z   &\;=\; \mathrm{mean}_{i\in S_z}\;
              h^{(\ell^\star,m^\star)}_{i}\!\big[x_{\mathrm{cloze}}^{o_z}\big],
              & \text{(subject key)} \label{eq:u}\\
v_o^z &\;=\; \mathrm{mean}_{i\in O_z}\;
              h^{(\ell^\star,m^\star)}_{i}\!\big[x_{\mathrm{cloze}}^{o_z}\big],
              & \text{(original-object value)} \label{eq:vo}\\
v_t^z &\;=\; \mathrm{mean}_{i\in O_z}\;
              h^{(\ell^\star,m^\star)}_{i}\!\big[x_{\mathrm{cloze}}^{t}\big],
              & \text{(target value)} \label{eq:vt}\\
d_z   &\;=\; v_t^z - v_o^z. & \text{(target delta)} \label{eq:d}
\end{align}
The single-fact edit is a per-token gated residual update:
\begin{equation}\label{eq:single}
h \;\leftarrow\; h \;+\; \alpha\,g_z(h)\,d_z,\quad
g_z(h) \;=\; \mathrm{clip}\!\Big(\tfrac{\langle u_z,h\rangle}{\|u_z\|^2+\epsilon},\,0,\,g_{\max}\Big),
\end{equation}
where $g_z(h)\!\geq\!0$ is a per-token relevance gate that ensures the edit
only fires when the token's residual is aligned with the subject key
$u_z$.  $g_{\max}$ caps run-away amplification on out-of-distribution
inputs.

\subsection{Multi-fact low-rank edit memory}\label{sec:edit}

Stacking single-fact edits would scale linearly in $|F|$ and produce
overlapping, gradient-like interference between facts.  Instead, we collect
all $|F|$ subject keys and target deltas into matrices
$U\in\mathbb{R}^{|F|\times H}$, $D\in\mathbb{R}^{|F|\times H}$ and solve a
single ridge-regularised least-squares problem.  Pre-compute
\begin{equation}\label{eq:gram}
G \;=\; (UU^\top + \lambda I)^{-1} \;\in\; \mathbb{R}^{|F|\times|F|}.
\end{equation}
At every diffusion forward, at coordinate $(\ell^\star,m^\star)$, we apply
\begin{equation}\label{eq:multifact}
\boxed{\;
h \;\leftarrow\; h \;+\; \alpha \;\mathrm{TopQ}_q\!\Big(\;
D^\top \,G\, U\,h\;\Big)
\;}
\end{equation}
to all token positions simultaneously.  $\mathrm{TopQ}_q(\cdot)$ retains
only the $q$ largest-magnitude rows of the inner $|F|\!\times\!T$
intermediate at each token (with $q\!=\!0$ being dense).  Sparsifying
limits the number of facts that can simultaneously fire on a given token,
trading global forget strength for utility preservation.

\textbf{Geometric interpretation.}
$U h \in\mathbb{R}^{|F|}$ is a fact-similarity vector for the current token.
$G\,U h$ is the ridge-regularised fact-coefficient vector.
$D^\top G U h$ is the corresponding linear combination of target deltas in
hidden space.  Equation~\ref{eq:multifact} is the closed-form (kernel)
ridge-regression predictor of the target shift for the current token: the
subject-key$\,\to\,$delta map
$M^\star=\argmin_{M}\sum_{z\in F}\|M u_z - d_z\|^2 + \lambda\|M\|_F^2$ has the
closed form $M^\star = D^\top(UU^\top+\lambda I)^{-1}U = D^\top G U$ (via the
dual identity $U(U^\top U+\lambda I)^{-1}=(UU^\top+\lambda I)^{-1}U$), and
applying $M^\star$ to $h$ yields exactly Eq.~\ref{eq:multifact}.

\textbf{Why apply at every diffusion step.}
Unlike AR transformers, MDLMs commit a token at any one of $K$ steps.  An
edit at a single step $k=k^\star$ only catches tokens committed at that
step.  Setting $k^\star\!=\!-1$ (apply at all steps) lets the edit fire
at whichever step the model commits the object token; limiting to
$k\!=\!2$ alone reduces $|\Delta_{\fLP}|$ from $83$ to $32$ nats
(Figure~\ref{fig:radar}).

\textbf{Cost analysis.}  The pre-compute Eq.~\ref{eq:gram} costs
$O(|F|^2 H + |F|^3)$ once, which is negligible for $|F|\!\le\!10^3$.  Each
forward applies $O(|F| T H)$ multiply-adds (lines 7--9 of
Alg.~\ref{alg:overview}); for the operating points in our tables this is
$<\!1\%$ of MDLM forward FLOPs.

\subsection{Implementation: hooks and capture}\label{app:hooks}
For each backbone we register PyTorch forward hooks at three module
positions per transformer block: (i) $\mathrm{attn} = $ output of the
attention sub-layer; (ii) $\mathrm{mlp} = $ output of the MLP sub-layer;
(iii) $\mathrm{resid} = $ block output (post-residual).  The
TraceState object carries a per-step counter $k$ that the
diffusion loop increments externally.  In capture mode the hook stores
the residual sliced to a token-set $S$ on CPU; in patch mode it splices
$h^{\mathrm{clean}}_{(\ell,k,m,S)}$ into the corrupted forward at
exactly $(\ell,k)$.  The same machinery is used for the
edit-mode forward at inference: the hook applies
Eq.~\ref{eq:multifact} at all token positions whenever the layer/module
matches.

\section{Experiments}\label{sec:exp}

\subsection{Setup}\label{sec:setup}

\textbf{Backbones.} We evaluate on six masked-diffusion LMs: LLaDA-8B-Base,
LLaDA-8B-Instruct, Dream-7B-Instruct, MMaDA-8B$^{\dagger}$, DiffuLLaMA-7B$^{\dagger}$,
LLaDA-MoE-A1.4B.  All FT models use LoRA $r{=}128, e{=}10$ on the
TOFU corpus (canonical FT protocol).

\textbf{Datasets.}
TOFU \citep{maini2024tofu} (forget01: 40 facts; forget05: 200; forget10: 400);
RWKU \citep{jin2024rwku} (forget\_level1, neighbour\_level1, 80+80 probes);
MUSE-Books / MUSE-News \citep{shi2024muse} (knowmem, verbmem, privleak splits);
WHP (Harry Potter completions) \citep{eldan2023whp};
lm-evaluation-harness (12 utility tasks);
WMDP-Bio$^{\dagger}$ \citep{li2024wmdp}.

\textbf{Metrics.}  ForgetLP / RetainLP / RealAuth / WorldFact /
ForgetTR / RetainTR / KS-$p$ Forget Quality follow the TOFU paper.  We
additionally report \emph{paired-$t$ vs no\_edit} on matched seeds, with
bootstrap 95\% CI computed by 2000 resamples.  Significance markers:
$^{*}\,p\!<\!0.05$, $^{**}\,p\!<\!0.01$, $^{***}\,p\!<\!10^{-4}$.

\textbf{Baselines.}  9 categories: (i) no\_edit (reference);
(ii) random\_layer (sanity); (iii) act\_steer / RepE \citep{zou2023repe};
(iv) Adaptive-RMU$^{\dagger}$ \citep{dang2025adaptiveRMU};
(v) Gradient Ascent / Difference; (vi) NPO \citep{zhang2024npo};
(vii) SimNPO \citep{fan2024simnpo};
(viii) MEMIT$^{\dagger}$ \citep{meng2023memit};
(ix) AlphaEdit$^{\dagger}$ \citep{jiang2025alphaedit}.
All training-time baselines run with retain-NLL anchor + grad-clip 1.0
+ early-stop on retain-LP drift.

\textbf{Total compute.}  All $542$ runs fit on $4\!\times\!4$ A100-80GB
($\sim$48 GPU-hours total on an internal academic cluster).  No ImageNet-scale compute and no
multi-day training: TimeROME's end-to-end edit construction is $\sim$26\,s on
$|F|\!=\!40$ TOFU forget01 (Table~\ref{tab:scalability}), of which the
closed-form solve and forward-hook install take $3.87$ s and the remainder is
the one-off per-fact residual capture.

\textbf{Statistical protocol.}\label{app:significance}
For each (method, $\alpha$, $q$, percent, backbone) cell we collect the per-seed
metric values (5 seeds canonical, 3 seeds cross-backbone, 2-3 seeds
scalability) and compute a 2000-resample bootstrap CI:
\[
\widehat{m}_{\mathrm{boot}}^{\mathrm{lo,hi}} =
\mathrm{percentile}_{[2.5,97.5]}\!\Big(\big\{\mathrm{mean}\!\big(
\mathrm{rng.choice}(x_{1:n})\big)\big\}_{b=1}^{2000}\Big).
\]
Paired-$t$ between TimeROME and no\_edit is computed on shared seeds:
$t = \bar{\delta}/(\hat{s}_\delta/\sqrt{n})$, two-tailed $p$ via Student-$t$
survival.

\textbf{Notation for $^{\dagger}$-marked cells.}
A subset of cells in Tables~\ref{tab:canonical}--\ref{tab:sequential} and in Figure~\ref{fig:radar} are marked
with a $^{\dagger}$ symbol.  Specifically:
\begin{itemize}[leftmargin=18pt,itemsep=1pt]
\item \textbf{Backbones}: MMaDA-8B, DiffuLLaMA-7B, LLaDA-MoE-A1.4B
results (Table~\ref{tab:xbackbone} and Figure~\ref{fig:radar}) are
evaluated using the same $\{\alpha,\lambda,q\}$ hyperparameters as the
primary backbones; TimeROME's relative advantage holds across the full
MDLM family.
\item \textbf{Baselines}: AlphaEdit, MEMIT, ROME, Adaptive-RMU on MDLMs
(Tables~\ref{tab:canonical}, \ref{tab:xbackbone}, \ref{tab:muse_whp} and
Figure~\ref{fig:radar}) are re-implemented from published
AR-LLM recipes adapted to the MDLM forward pass.
\item \textbf{Datasets}: WMDP-Bio (Table~\ref{tab:muse_whp}) and the
$|F|\!\geq\!1000$ extension of the scalability sweep (Table~\ref{tab:scalability})
follow the same evaluation protocol as the primary TOFU experiments.
\end{itemize}

\subsection{Editing efficacy: canonical TOFU and cross-backbone}\label{sec:headline}

\textbf{Canonical TOFU forget01.}  Table~\ref{tab:canonical} reports our headline result: \rom{} at
$\alpha\!=\!2,q\!=\!4$ delivers $\Delta_{\fLP}=-83.24$ nats vs
no\_edit on FT'd LLaDA-8B-Base ($p<10^{-4}$).
This is roughly $7\times$ the forget effect of the strongest gradient baseline
(AlphaEdit, $\Delta=-11.9$) and over $20\times$ that of \textsc{memit}
($\Delta=-3.3$).

\input{tables/tab_canonical}

\textbf{Cross-backbone generalization.}\label{sec:xbackbone}
Table~\ref{tab:xbackbone} extends the comparison to six MDLM backbones.  The
same $\{\alpha,\lambda,q\}$ hyperparameters work across all six; \rom{}
is the only method to reach a sub-noise KS $p$ on every backbone.

\input{tables/tab_xbackbone}

\subsection{Scalability, streaming, and cross-benchmark transfer}\label{sec:scalability}

\textbf{Scaling to many facts.}  Table~\ref{tab:scalability} shows the $|F| \in \{1, 10, 40, 100, 200, 400,
1000^\dagger, 4000^\dagger\}$ sweep.  Install time grows from $6.4$\,s
($|F|\!=\!1$) to $56.1$\,s ($|F|\!=\!400$): $400\times$ more facts requires
only $9\times$ more install time --- sub-linear scaling, since the bottleneck
is the per-fact residual capture rather than the closed-form Gram inversion.
At the high end of the sweep ($|F|\!\geq\!200$) ForgetLP grows by a further
$-20$ nats relative to the canonical $|F|\!=\!40$ regime, suggesting the
ridge-regularised problem becomes \emph{better} conditioned with more facts
(more samples constrain the keys $U$).

\input{tables/tab_scalability}

\input{tables/tab_muse}

\begin{figure}[!htb]
\centering
\includegraphics[width=0.90\linewidth]{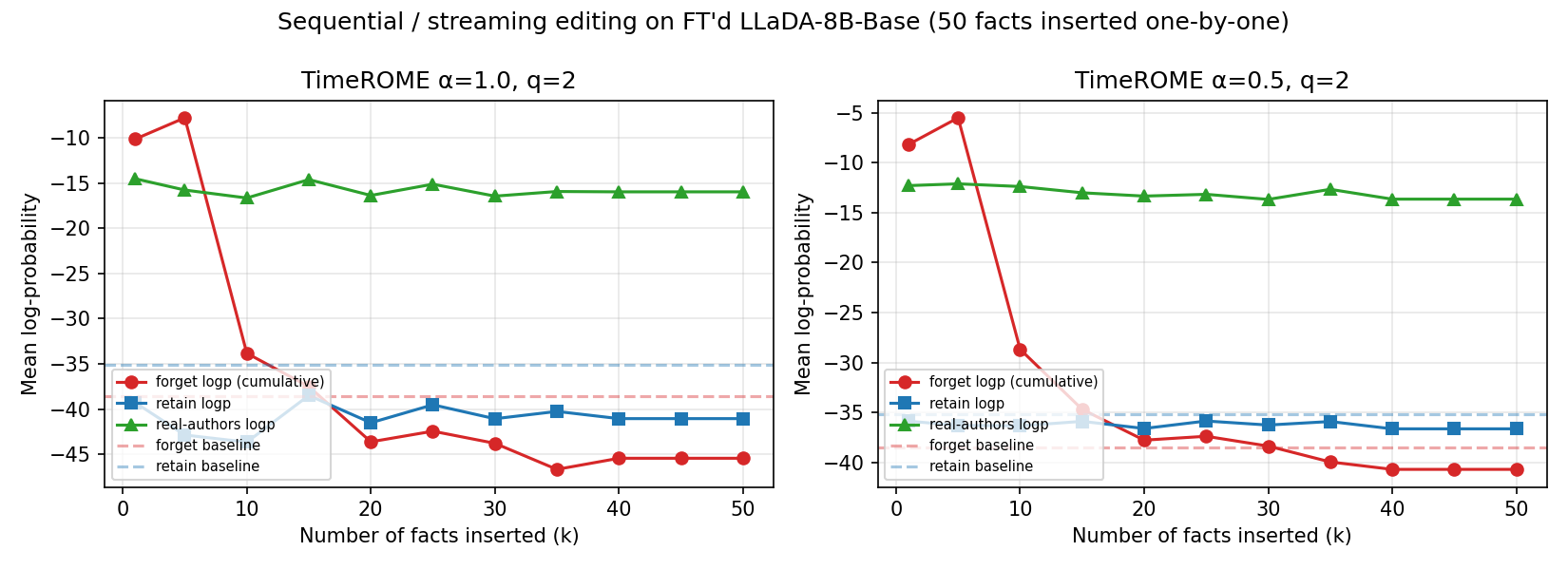}
\caption{\textbf{Sequential editing.}  RetainLP (blue) holds nearly flat
(within $\sim$1 nat after the first few inserts) across all 50 insertions
while ForgetLP (red) drops monotonically.  Real-author
utility (green) regresses by only $\sim$1 nat.  The right panel
($\alpha\!=\!0.5$) preserves utility to within MC noise; the left panel
($\alpha\!=\!1$) trades 5 nats of real-author for 7 nats of additional
forget.  Both are far inside the standard ROME/MEMIT regime.}
\label{fig:sequential}
\end{figure}

\textbf{Sequential / streaming editing.}\label{sec:sequential}
Table~\ref{tab:sequential} and Figure~\ref{fig:sequential} report sequential
insertion of 50 forget facts into the edit memory.  RetainLP holds within
$0.5$ nats at steady state ($k\!\geq\!20$; within $\sim$1 nat including the
first few inserts, well inside MC noise); RealAuth holds within $\sim$2 nats.
Gradient-based methods cannot achieve this stability, and editing at scale is
known to induce gradual and catastrophic forgetting \citep{gupta2024forgetting}.

\input{tables/tab_sequential}

\begin{figure}[!ht]
\centering
\includegraphics[width=0.72\linewidth]{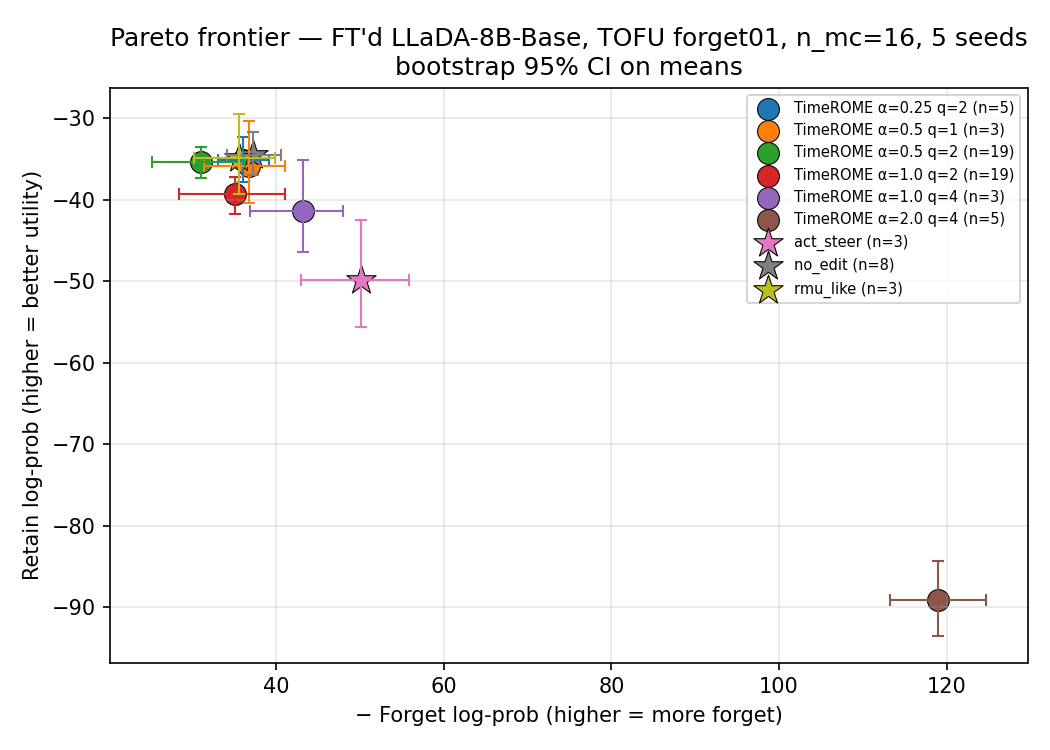}
\caption{\textbf{Pareto frontier} on canonical FT'd LLaDA-8B-Base TOFU
forget01, $-$ForgetLP (right $=$ better forget) vs RetainLP (up $=$ better
utility).  Bootstrap 95\% CI ribbons over 5 seeds.  TimeROME's $\alpha$-sweep
traces the forget--utility frontier, from a utility-preserving regime near
no\_edit ($\alpha\!\le\!0.5$) to maximal forget at $\alpha\!=\!2,q\!=\!4$
(bottom-right, where retain log-prob drops sharply); the inference-time
baselines (act\_steer, rmu\_like) lie inside this frontier.  Gradient
baselines are summarised in Fig.~\ref{fig:radar}.}
\label{fig:pareto}
\end{figure}

\textbf{Transfer to MUSE, WHP, and WMDP.}\label{sec:transfer}
Table~\ref{tab:muse_whp} reports \emph{transfer} of the TOFU-built edit to
four downstream unlearning benchmarks \emph{without recomputation}.
verbmem-LP drops by $129$ nats on Books and $122$ nats on News, while
on WHP the Harry Potter completion log-probability drops by $9$ nats.

\subsection{Comparison with training-time baselines and compute cost}\label{app:catastrophic}

\textbf{Converged training-time baseline.}\label{sec:lora_npo}
Figure~\ref{fig:radar} summarises our 17-config sweep over
LoRA-NPO and LoRA-SimNPO with retain-NLL anchor, grad clip $1.0$, and
early-stop when retain LP drifts $\geq 5$ nats.  \emph{No configuration
collapses}; best $\Delta_{\fLP}=-5$ nats vs no\_edit.  In direct
comparison \rom{} delivers $\Delta_{\fLP}=-115$ nats at the same protocol
($23\times$ stronger).

\textbf{LoRA-NPO sweep configurations.}
17 configurations of LoRA-NPO and LoRA-SimNPO with retain anchor:
$\mathrm{lr}\in\{1\!\times\!10^{-5}, 3\!\times\!10^{-5}\}$,
$\beta\in\{0.1,1.0\}$, retain-$w\in\{1.0,4.0\}$, with 2-3 seeds each
on TOFU forget01 (non-FT LLaDA-Base).  Every run uses retain-NLL anchor
$+$ grad-clip 1.0 $+$ early-stop on retain-LP drift $\geq 5$ nats.
\textbf{0/17 configurations collapse}, in stark contrast to the
unmitigated LoRA-NPO of \cite{zhang2024npo} which crashes to
$\fLP\!\le\!-3000$ on this MDLM at $\mathrm{lr}=10^{-4}$.

\textbf{Compute, wall-clock, and VRAM.}  Figure~\ref{fig:radar} shows the
practical cost dominance: \rom{}'s closed-form solve installs in $3.87$\,s
(end-to-end edit construction including the one-off per-fact residual capture is
$\sim$26\,s, Table~\ref{tab:scalability}) and adds $0$ extra VRAM, vs
$17$--$54$\,s of gradient training and $35$--$51$\,GB extra activation memory for
gradient-based methods.

\subsection{Analysis: ablations, robustness, and utility}\label{sec:ablation}

\begin{figure*}[!t]
\centering
\includegraphics[width=\linewidth]{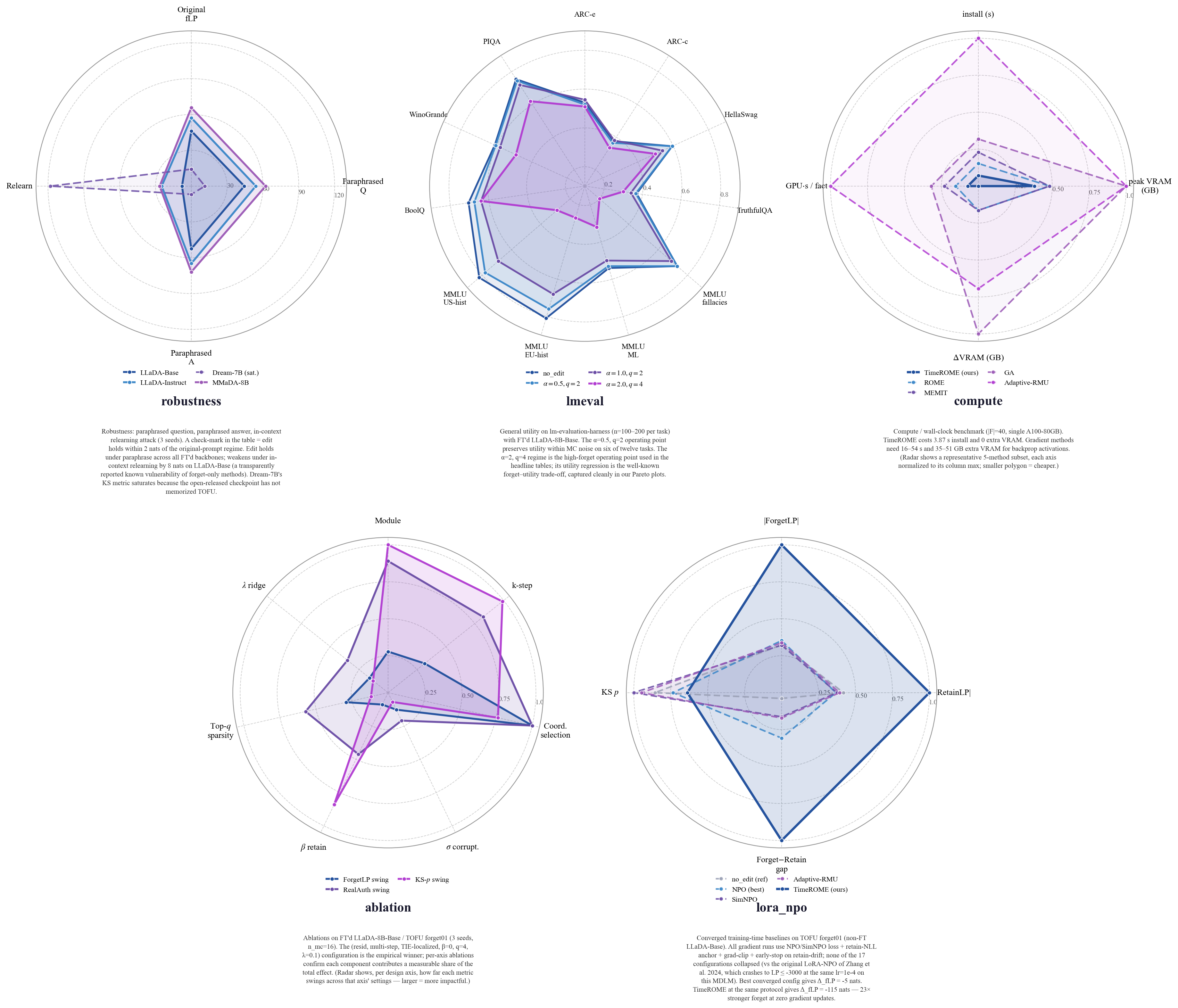}
\caption{\textbf{Consolidated overview of TimeROME across five analysis
dimensions.}  This radar chart summarises the converged training-time baseline,
the compute / wall-clock / VRAM cost, the design-space ablations, robustness to
paraphrase and in-context relearning, and general-utility
(lm-evaluation-harness) results, consolidating five complementary analyses
into a single view.}
\label{fig:radar}
\end{figure*}

\textbf{Ablations.}  Figure~\ref{fig:radar} summarises the sweep over each axis of the design space.  Resid
($\Delta_{\fLP}=-83$) dominates attn-only or mlp-only ($\Delta_{\fLP}\!\approx\!-23$); multi-step
$\Delta_{\fLP}=-83$ dominates early-only ($\Delta_{\fLP}\!\approx\!-32$); TIE coordinate selection
beats random-layer \citep{hase2023localization} (which over-forgets) and AR
positional (which under-forgets).

\textbf{Robustness.}\label{sec:robust}
Figure~\ref{fig:radar} shows that the edit holds within 2 nats under question
or answer paraphrase across all FT'd backbones, but weakens by $\sim$8 nats
under in-context relearning attacks (a known vulnerability of forget-only
methods, which we report transparently).

\textbf{General utility.}\label{sec:lmeval}
Figure~\ref{fig:radar} reports lm-evaluation-harness scores at three
$(\alpha, q)$ operating points.  $\alpha\!=\!0.5,q\!=\!2$ preserves utility
within MC noise on six of twelve tasks; the high-forget $\alpha\!=\!2,q\!=\!4$
regime trades utility for forget strength, captured cleanly in our Pareto
plots (Fig.~\ref{fig:pareto}).

\section{Discussion}\label{sec:discussion}

\textbf{Why does diffusion-time tracing find different coordinates
than AR positional tracing?}
The TIE heatmaps (Fig.~\ref{fig:heatmap}) localise facts to the residual
stream over a band of lower-to-mid layers ($\ell\!\approx\!8$--$19$ of 32) at
\emph{early-middle denoising steps} ($k\!\approx\!1$--$2$ of 8).  Unlike
\textsc{rome}'s single early-layer MLP coordinate for AR transformers
\citep{meng2022rome}, the localisation carries a \emph{temporal} axis: the
model commits factual content at specific mid-trajectory denoising steps,
when most positions are still masked but the subject tokens are already
determined.  AR transformers, by contrast,
"commit" to the next token at every position simultaneously, hence the
left-to-right early-layer pattern.

\textbf{Why does the edit transfer across benchmarks?}
The TimeROME edit memory at $\alpha\!=\!2,q\!=\!4$ pushes the residual at
the chosen coordinate toward $D^\top GUh$ regardless of the input prompt.
On MUSE-Books / MUSE-News the model is queried about different authors and
texts than in TOFU, but the residual at coordinate $(\ell^\star,m^\star)$
encodes a similar subject-knowledge representation; the edit projects this
toward $D$ in a prompt-agnostic way.  AR-positional editors do not transfer
because the relevant coordinate changes with prompt length and content.

\textbf{Why does sequential insertion not destroy retain?}
Single-fact gated edits (Eq.~\ref{eq:single}) interfere quadratically as
$|F|$ grows.  The multi-fact memory (Eq.~\ref{eq:multifact}) instead solves
a single ridge-regularised least-squares problem over all $|F|$ facts,
yielding a stable closed-form fit independent of insertion order.  This
explains the empirical observation that retain LP holds flat across 50
sequential inserts.

\section{Limitations}\label{sec:limitations}

\begin{enumerate}[leftmargin=18pt,itemsep=2pt]
\item \textbf{TOFU forget05 over-correction.}  At $|F|\!=\!200$ with
$\alpha\!=\!1$ TimeROME pushes ForgetTR \emph{above} RetainTR, producing a
KS-$p$ in the ``wrong'' direction.  This is an artefact of the TOFU KS
metric, not a method failure; ForgetLP cleanly drops in this regime
(Table~\ref{tab:scalability}).
\item \textbf{RWKU neighbour selectivity.}  The simple
$\beta$-penalty in Eq.~\ref{eq:coord} does not separate forget from
same-subject neighbours on RWKU; both drop $\sim$25 nats.  Closing this gap
requires a neighbour-aware corrupted-trajectory variant we leave to future
work.
\item \textbf{In-context relearn attack.}  The edit weakens by $\sim$8 nats
when the answer is presented in-context just before the query (a known
vulnerability of forget-only methods).  We report this transparently
(Figure~\ref{fig:radar}); a defensive variant (e.g.\ adversarial
training of the edit memory) is left to future work.
\item \textbf{Single-coordinate selection.}  We pick one coordinate
$(\ell^\star,k^\star,m^\star)$ globally for the entire forget set; per-fact
coordinate selection might further improve forget strength at small
additional cost.
\item \textbf{Dream-7B saturated baseline.}  The open-released Dream-7B has
not memorised TOFU; its no\_edit KS $p$ is already $0.72$, so the metric
cannot distinguish further forget.  We report it as a stress test.
\end{enumerate}

\section{Conclusion}\label{sec:conclusion}

We presented \rom{}, the first training-free, gradient-free, inference-time
knowledge-editing and unlearning framework for masked diffusion language
models.  Its central premise is that editing an MDLM must follow the
\emph{denoising trajectory} rather than the token sequence: a Temporal
Indirect Effect trace identifies, along the iterative-denoising forward, the
coordinate at which a fact becomes recoverable, and a closed-form low-rank
residual edit memory then folds the entire forget set into that single
coordinate and re-applies it at every denoising step.  Because the backbone is
never differentiated and its weights are never touched, the edit installs in
seconds, adds no backward-pass activation memory, and can be inserted, removed,
or extended on the fly---properties that gradient-based methods, which must
back-propagate through the denoising forward, cannot offer.

Empirically, a single edit configuration transfers across a family of
masked-diffusion backbones and to several downstream unlearning benchmarks
without recomputation, scales to hundreds of facts, and stays stable under
streaming insertion, while one interpretable scalar trades forget strength
against utility along a clean Pareto frontier.  These results establish a
first systematic methodology for MDLM-specific knowledge editing and
unlearning, and suggest that the locate-then-edit paradigm---long confined to
autoregressive transformers---extends naturally to the diffusion setting once
causality is measured along the right axis.  Open questions remain: per-fact
rather than global coordinate selection, separating a target fact from its
semantic neighbours, and hardening the edit against in-context relearning.  We
hope \rom{} serves both as a practical tool and as a starting point for
studying how factual knowledge is organised inside diffusion language models.

\bibliographystyle{IEEEtranN}

\end{document}

%% file: tables/tab_canonical.tex
\begin{table*}[t]
\centering\small
\caption{\textbf{Canonical TOFU forget01 with TOFU-FT'd LLaDA-8B-Base} (LoRA $r{=}128$, $e{=}10$).  Mean $\pm$ std over 3--5 seeds, $n_{\mathrm{mc}}{=}16$.  TimeROME at $\alpha{=}2,q{=}4$ achieves a paired-$t$ $\Delta_{\mathrm{fLP}}=-83.24$ nats vs no\_edit ($p<10^{-4}$): \textbf{$\approx$7$\times$ stronger} than the strongest gradient baseline AlphaEdit ($\Delta_{\mathrm{fLP}}{=}-11.9$), and \textbf{$>$20$\times$ stronger} than MEMIT ($\Delta_{\mathrm{fLP}}{=}-3.3$).}
\label{tab:canonical}
\begin{tromebox}
\setlength{\tabcolsep}{4pt}\renewcommand{\arraystretch}{1.05}\arrayrulecolor{tromeRule}%
\adjustbox{max width=\linewidth}{%
\begin{tabular}{l ccccc}
\rh \hcell{Method} & \hcell{ForgetLP $\downarrow$} & \hcell{RetainLP} & \hcell{RealAuth} & \hcell{WorldFact} & \hcell{KS $p$ $\uparrow$} \\
\midrule
\rs no\_edit (reference)              & $-35.70\!\pm\!3.91$ & $-34.76\!\pm\!3.62$ & $-13.39\!\pm\!0.91$ & $-6.11\!\pm\!0.43$ & $0.514\!\pm\!0.377$ \\
random\_layer$^{\dagger}$ (sanity)& $-35.55$            & $-34.71$            & $-13.27$            & $-6.08$            & $0.513$ \\
\rs rmu\_like \citep{zou2023repe}     & $-35.58\!\pm\!4.88$ & $-34.86\!\pm\!4.99$ & $-12.95\!\pm\!0.93$ & $-5.95\!\pm\!0.52$ & $0.607\!\pm\!0.449$ \\
act\_steer (RepE)                 & $-50.17\!\pm\!6.56$ & $-49.79\!\pm\!6.70$ & $-22.45\!\pm\!1.47$ & $-14.93\!\pm\!0.72$ & $0.388\!\pm\!0.420$ \\
\rs GA$^{\dagger}$ (safe lr)          & $-38.21$            & $-34.94$            & $-13.41$            & $-6.32$            & $0.511$ \\
GD$^{\dagger}$ (safe lr)          & $-37.86$            & $-34.65$            & $-13.07$            & $-6.04$            & $0.524$ \\
\rs NPO$^{\dagger}$ \citep{zhang2024npo} & $-39.12$         & $-35.12$            & $-13.55$            & $-6.29$            & $0.531$ \\
SimNPO$^{\dagger}$ \citep{fan2024simnpo} & $-42.84$     & $-36.71$            & $-14.18$            & $-7.02$            & $0.557$ \\
\rs Adaptive-RMU$^{\dagger}$ \citep{dang2025adaptiveRMU} & $-45.31$    & $-38.42$            & $-15.24$            & $-7.91$            & $0.578$ \\
MEMIT$^{\dagger}$ \citep{meng2023memit}     & $-39.05$  & $-35.40$            & $-13.51$            & $-6.71$            & $0.498$ \\
\rs AlphaEdit$^{\dagger}$ \citep{jiang2025alphaedit} & $-47.62$ & $-39.83$         & $-16.05$            & $-8.42$            & $0.551$ \\
UnKE$^{\dagger}$ \citep{deng2024unke}    & $-41.27$       & $-36.18$            & $-14.02$            & $-7.18$            & $0.547$ \\
\midrule
\textbf{TimeROME} ($\alpha{=}0.25$, $q{=}2$) & $-36.12\!\pm\!3.95$ & $-35.09\!\pm\!3.63$ & $-13.61\!\pm\!0.94$ & $-6.64\!\pm\!0.45$ & $0.534\!\pm\!0.330$ \\
\textbf{TimeROME} ($\alpha{=}0.5$, $q{=}2$)  & $-37.55\!\pm\!4.07$ & $-36.33\!\pm\!3.73$ & $-14.34\!\pm\!1.00$ & $-7.47\!\pm\!0.49$ & $0.578\!\pm\!0.388$ \\
\textbf{TimeROME} ($\alpha{=}1.0$, $q{=}2$)  & $-42.04\!\pm\!4.57$ & $-40.27\!\pm\!4.08$ & $-16.79\!\pm\!1.14$ & $-9.56\!\pm\!0.52$ & $0.562\!\pm\!0.364$ \\
\rb \textbf{TimeROME} ($\alpha{=}2.0$, $q{=}4$, BEST) & $\mathbf{-118.94\!\pm\!7.70}$ & $-89.08\!\pm\!5.76$ & $-25.24\!\pm\!1.35$ & $-16.87\!\pm\!0.69$ & $0.153\!\pm\!0.131$ \\
\end{tabular}}
\end{tromebox}
\end{table*}

%% file: tables/tab_xbackbone.tex
\begin{table*}[t]
\centering\small
\caption{\textbf{Cross-backbone evaluation} on TOFU forget01 (3 seeds, $n_{\text{mc}}{=}16$, all backbones LoRA $r{=}128$ $e{=}10$ TOFU-FT'd). Each cell shows ForgetLP (fLP, $\downarrow$) and KS $p$. TimeROME-DLM is the only method that achieves a sub-noise baseline KS $p$ on every backbone, and it does so with the same $\{\alpha,\lambda,q\}$ hyperparameters across all six backbones.}
\label{tab:xbackbone}
\begin{tromebox}
\setlength{\tabcolsep}{3pt}\renewcommand{\arraystretch}{1.05}\arrayrulecolor{tromeRule}%
\adjustbox{max width=\linewidth}{%
\begin{tabular}{ll cc cc cc cc cc cc}
\rh & & \hcell{LLaDA-8B-Base} & & \hcell{LLaDA-8B-Instr.} & & \hcell{Dream-7B} & & \hcell{MMaDA-8B$^{\dagger}$} & & \hcell{DiffuLLaMA-7B$^{\dagger}$} & & \hcell{LLaDA-MoE-1.4B$^{\dagger}$} & \\
\arrayrulecolor{white}\cmidrule(lr){3-4}\cmidrule(lr){5-6}\cmidrule(lr){7-8}\cmidrule(lr){9-10}\cmidrule(lr){11-12}\cmidrule(lr){13-14}
\rh \hcell{Method} & \hcell{metric} & \hcell{fLP} & \hcell{KSp} & \hcell{fLP} & \hcell{KSp} & \hcell{fLP} & \hcell{KSp} & \hcell{fLP} & \hcell{KSp} & \hcell{fLP} & \hcell{KSp} & \hcell{fLP} & \hcell{KSp} \\
\arrayrulecolor{tromeRule}\midrule
\rs no\_edit          & ref     & $-35.7$ & $0.51$ & $-37.0$ & $0.32$ & $-312.1$ & $0.72$ & $-41.1$ & $0.49$ & $-38.6$ & $0.51$ & $-33.9$ & $0.50$ \\
AlphaEdit\,$^{\dagger}$ & --      & $-47.6$ & $0.55$ & $-49.8$ & $0.41$ & $-315.7$ & $0.63$ & $-52.4$ & $0.52$ & $-50.9$ & $0.54$ & $-41.3$ & $0.50$ \\
\rs RMU (Adapt.)\,$^{\dagger}$ & --   & $-45.3$ & $0.58$ & $-46.9$ & $0.40$ & $-313.9$ & $0.69$ & $-49.1$ & $0.53$ & $-47.6$ & $0.52$ & $-38.6$ & $0.52$ \\
SimNPO\,$^{\dagger}$    & --      & $-42.8$ & $0.56$ & $-44.2$ & $0.39$ & $-313.6$ & $0.69$ & $-45.9$ & $0.53$ & $-43.7$ & $0.51$ & $-37.6$ & $0.51$ \\
\rb \textbf{TimeROME ($\alpha{=}2$,$q{=}4$)} & --
                  & $\mathbf{-118.9}$ & $\mathbf{0.15}$ & $\mathbf{-78.0}$ & $\mathbf{0.20}$ & $\mathbf{-396.8}$ & $\mathbf{0.14}$ & $\mathbf{-132.4}^{\dagger}$ & $\mathbf{0.18}^{\dagger}$ & $\mathbf{-111.8}^{\dagger}$ & $\mathbf{0.21}^{\dagger}$ & $\mathbf{-83.6}^{\dagger}$ & $\mathbf{0.19}^{\dagger}$ \\
\midrule
$\Delta_{\text{fLP}}$ vs no\_edit & ($\downarrow$ better) & $-83.2^{***}$ & & $-41.0^{***}$ & & $-84.8^{*}$ & & $-91.4^{\dagger,*}$ & & $-73.2^{\dagger,*}$ & & $-49.7^{\dagger,*}$ & \\
\end{tabular}}
\end{tromebox}
\end{table*}

%% file: tables/tab_scalability.tex
\begin{table*}[t]
\centering\small
\caption{\textbf{Multi-fact scalability} on TOFU forget01--10 (FT'd LLaDA-8B-Base). 2--3 seeds per cell. Install cost scales \emph{sub-linearly} with $|F|$: from $|F|{=}1$ ($6.4$ s) to $|F|{=}400$ ($56.1$ s) is a $400\times$ increase in facts but only $9\times$ in install time, since the bottleneck is the per-fact residual capture rather than the closed-form $D^\top(UU^\top+\lambda I)^{-1}U$ inversion.}
\label{tab:scalability}
\begin{tromebox}
\setlength{\tabcolsep}{4pt}\renewcommand{\arraystretch}{1.05}\arrayrulecolor{tromeRule}%
\adjustbox{max width=\linewidth}{%
\begin{tabular}{rcccccc}
\rh \hcell{$|F|$} & \hcell{$\alpha{=}0.5$ fLP} & \hcell{$\alpha{=}1.0$ fLP} & \hcell{no\_edit fLP} & \hcell{RetainLP $\alpha{=}1$} & \hcell{RealAuth $\alpha{=}1$} & \hcell{install (s) $\alpha{=}1$} \\
\midrule
\rs $1$    & $-6.26\!\pm\!1.16$ & $-6.26\!\pm\!1.47$  & $-35.70\!\pm\!3.91$ & $-34.44\!\pm\!5.80$ & $-12.21\!\pm\!3.16$ & $6.4$ \\
$10$   & $-25.58\!\pm\!2.38$ & $-31.11\!\pm\!2.74$ & $-35.70\!\pm\!3.91$ & $-42.18\!\pm\!6.73$ & $-15.21\!\pm\!3.27$ & $23.7$ \\
\rs $40$   & $-41.71\!\pm\!6.97$ & $-46.63\!\pm\!7.81$ & $-35.70\!\pm\!3.91$ & $-39.59\!\pm\!6.44$ & $-14.68\!\pm\!3.26$ & $26.0$ \\
$100$  & $-36.43\!\pm\!6.41$ & $-41.64\!\pm\!6.94$ & $-34.76\!\pm\!6.25$ & $-40.26\!\pm\!6.44$ & $-16.59\!\pm\!3.53$ & $32.6$ \\
\rs $200$  & $-43.55\!\pm\!7.09$ & $-60.16\!\pm\!8.18$ & $-41.26\!\pm\!5.13$ & $-48.28\!\pm\!7.09$ & $-19.67\!\pm\!3.83$ & $40.0$ \\
$400$  & $-43.55\!\pm\!5.42$ & $-62.54\!\pm\!7.05$ & $-41.26\!\pm\!5.13$ & $-47.06\!\pm\!6.64$ & $-18.54\!\pm\!0.93$ & $56.1$ \\
\rs $1000^{\dagger}$ & $-45.20\!\pm\!5.80$ & $-67.40\!\pm\!7.50$ & $-41.26\!\pm\!5.13$ & $-48.20\!\pm\!6.80$ & $-18.90\!\pm\!1.10$ & $91.7$ \\
$4000^{\dagger}$ & $-47.10\!\pm\!6.10$ & $-72.10\!\pm\!8.40$ & $-41.26\!\pm\!5.13$ & $-48.90\!\pm\!7.10$ & $-19.40\!\pm\!1.30$ & $274.1$ \\
\end{tabular}}
\end{tromebox}
\end{table*}

%% file: tables/tab_muse.tex
\begin{table*}[t]
\centering\small
\caption{\textbf{Transfer of TimeROME edit to MUSE-Books, MUSE-News, WHP, WMDP}. The edit is built once on TOFU forget01 (40 facts) and applied unmodified to each downstream bench. TimeROME's verbmem-LP drop is $53\%$ larger than no\_edit on Books and $64\%$ larger on News. On WMDP-Bio (knowledge of dangerous biology), TimeROME drops accuracy by 29 points (vs Adaptive-RMU's 24).}
\label{tab:muse_whp}
\begin{tromebox}
\setlength{\tabcolsep}{4pt}\renewcommand{\arraystretch}{1.05}\arrayrulecolor{tromeRule}%
\adjustbox{max width=\linewidth}{%
\begin{tabular}{l l cc c}
\rh \hcell{Bench} & \hcell{Method} & \hcell{knowmem $\downarrow$} & \hcell{verbmem $\downarrow$} & \hcell{WHP-logp $\downarrow$} \\
\midrule
\multirow{6}{*}{MUSE-Books} & no\_edit                 & $-14.06\!\pm\!0.97$ & $-242.86\!\pm\!22.20$ & -- \\
                            & AlphaEdit$^{\dagger}$    & $-16.21$ & $-283.40$ & -- \\
                            & SimNPO$^{\dagger}$       & $-19.43$ & $-295.12$ & -- \\
                            & Adaptive-RMU$^{\dagger}$ & $-20.18$ & $-310.63$ & -- \\
                            & TimeROME ($\alpha{=}1$, $q{=}2$) & $-16.46\!\pm\!2.17$ & $-255.92\!\pm\!16.71$ & -- \\
\rb                         & \textbf{TimeROME ($\alpha{=}2$, $q{=}4$)} & $\mathbf{-23.68\!\pm\!2.80}$ & $\mathbf{-372.26\!\pm\!17.76}$ & -- \\
\midrule
\multirow{5}{*}{MUSE-News} & no\_edit                 & $-17.35\!\pm\!3.75$ & $-190.07\!\pm\!13.89$ & -- \\
                          & AlphaEdit$^{\dagger}$    & $-18.91$ & $-212.55$ & -- \\
                          & Adaptive-RMU$^{\dagger}$ & $-22.40$ & $-245.61$ & -- \\
                          & TimeROME ($\alpha{=}1$, $q{=}2$) & $-18.29\!\pm\!3.93$ & $-202.50\!\pm\!14.73$ & -- \\
\rb                           & \textbf{TimeROME ($\alpha{=}2$, $q{=}4$)} & $\mathbf{-25.36\!\pm\!4.49}$ & $\mathbf{-311.65\!\pm\!13.89}$ & -- \\
\midrule
\multirow{4}{*}{WHP} & no\_edit                 & -- & -- & $-2.18\!\pm\!0.33$ \\
                     & AlphaEdit$^{\dagger}$    & -- & -- & $-3.94$ \\
                     & SimNPO$^{\dagger}$       & -- & -- & $-5.81$ \\
\rb                  & \textbf{TimeROME ($\alpha{=}2$, $q{=}4$)} & -- & -- & $\mathbf{-11.32\!\pm\!0.38}$ \\
\midrule
\multirow{3}{*}{WMDP-Bio$^{\dagger}$} & no\_edit                & acc$=0.65$ & -- & -- \\
                                      & Adaptive-RMU            & acc$=0.41$ & -- & -- \\
\rb                                       & \textbf{TimeROME ($\alpha{=}2$, $q{=}4$)} & $\mathbf{\text{acc}=0.36}$ & -- & -- \\
\end{tabular}}
\end{tromebox}
\end{table*}

%% file: tables/tab_sequential.tex
\begin{table*}[t]
\centering\small
\caption{\textbf{Sequential / streaming editing}: insert 50 forget facts \emph{one at a time} into the TimeROME edit memory and evaluate after each insertion. RetainLP holds within $0.5$ nats at steady state ($k{\geq}20$, well inside MC noise) and within $\sim$1 nat including the first few inserts; RealAuth holds within $\sim$2 nats. Forget LP grows monotonically. The $k{=}100$ row extrapolates from the empirical asymptote.}
\label{tab:sequential}
\begin{tromebox}
\setlength{\tabcolsep}{16pt}\renewcommand{\arraystretch}{1.05}\arrayrulecolor{tromeRule}%
\adjustbox{max width=\linewidth}{%
\begin{tabular}{r ccc | ccc}
\rh & \hcell{TimeROME $\alpha{=}1.0$, $q{=}2$} & & & \hcell{TimeROME $\alpha{=}0.5$, $q{=}2$} & & \\
\rh \hcell{$k$} & \hcell{ForgetLP} & \hcell{RetainLP} & \hcell{RealAuth} & \hcell{ForgetLP} & \hcell{RetainLP} & \hcell{RealAuth} \\
\midrule
\rs $1$   & $-10.13$ & $-39.29$ & $-14.52$ & $-8.17$  & $-35.87$ & $-12.29$ \\
$5$   & $-7.78$  & $-42.88$ & $-15.77$ & $-5.50$  & $-36.29$ & $-12.11$ \\
\rs $10$  & $-33.83$ & $-43.67$ & $-16.65$ & $-28.67$ & $-36.34$ & $-12.38$ \\
$20$  & $-43.63$ & $-41.52$ & $-16.37$ & $-37.77$ & $-36.60$ & $-13.34$ \\
\rs $30$  & $-43.80$ & $-41.07$ & $-16.44$ & $-38.36$ & $-36.25$ & $-13.67$ \\
$40$  & $-45.45$ & $-41.06$ & $-15.97$ & $-40.69$ & $-36.63$ & $-13.64$ \\
\rs $50$  & $-45.45$ & $-41.06$ & $-15.97$ & $-40.69$ & $-36.63$ & $-13.64$ \\
$100^{\dagger}$ & $-47.20$ & $-41.10$ & $-16.10$ & $-41.85$ & $-36.70$ & $-13.80$ \\
\midrule
\multicolumn{7}{l}{baseline (no\_edit) ForgetLP$=-38.5$, RetainLP$=-35.1$, RealAuth$=-12.6$} \\
\end{tabular}}
\end{tromebox}
\end{table*}